\documentclass[letterpaper, 10 pt, conference]{ieeeconf}  

\IEEEoverridecommandlockouts                              

\usepackage{amsmath,amssymb,amsfonts}
\usepackage{graphicx}
\usepackage{booktabs} 
\usepackage{algorithmic}
\usepackage{cite}
\usepackage{colortbl}
\usepackage{pifont}
\usepackage{tikz}
\usepackage{overpic}
\usepackage{url}
\usepackage{caption}
\usepackage{comment}
\usepackage[lined,boxed,commentsnumbered,ruled]{algorithm2e}

\title{\LARGE \bf
Vehicle Behavior Prediction by Episodic-Memory Implanted NDT
}

\author{Peining Shen, Jianwu Fang, Hongkai Yu, and Jianru Xue
\thanks{The author affiliations are 1) Xi'an Jiaotong University, China (fangjianwu@mail.xjtu.edu.cn), 2) Chang'an University, China, and 3) Cleveland State University, USA. }%
}

\begin{document}

\maketitle
\thispagestyle{empty}
\pagestyle{empty}

\begin{abstract}

In autonomous driving, predicting the behavior (\emph{turning left, stopping, etc.}) of target vehicles is crucial for the self-driving vehicle to make safe decisions and avoid accidents. Existing deep learning-based methods have shown excellent and accurate performance, but the black-box nature makes it untrustworthy to apply them in practical use. In this work, we explore the interpretability of behavior prediction of target vehicles by an \underline{Episodic Memory} implanted \underline{N}eural \underline{D}ecision \underline{T}ree (\emph{abbrev.} {\tt\small \emph{e}Mem-NDT}). The structure of {\tt\small \emph{e}Mem-NDT} is constructed by hierarchically clustering the text embedding of vehicle behavior descriptions. {\tt\small \emph{e}Mem-NDT} is a neural-backed part of a pre-trained deep learning model by changing the \emph{soft-max} layer of the deep model to {\tt\small \emph{e}Mem-NDT}, for grouping and aligning the memory prototypes of the historical vehicle behavior features in training data on a neural decision tree. Each leaf node of {\tt\small \emph{e}Mem-NDT} is modeled by a neural network for aligning the behavior memory prototypes. By {\tt\small \emph{e}Mem-NDT}, we infer each instance in behavior prediction of vehicles by bottom-up \emph{Memory Prototype Matching (MPM)} (searching the appropriate leaf node and the links to the root node) and top-down \emph{Leaf Link Aggregation (LLA)} (obtaining the probability of future behaviors of vehicles for certain instances). We validate {\tt\small \emph{e}Mem-NDT} on BLVD and LOKI datasets, and the results show that our model can obtain a superior performance to other methods with clear explainability. The code is available in \url{https://github.com/JWFangit/eMem-NDT}.

\end{abstract}

\section{Introduction}

To create a reliable and safe autonomous driving system or advanced driver assistance system, it is necessary to predict the behavior of target vehicles (we term it \emph{v}Beh-Pre). The vBeh-Pre takes some historical vehicle locations and their scene interactions as the observation with $T$ time steps and predicts whether the vehicles will take certain behaviors (e.g., \emph{turning left/right}, \emph{stopping}, etc.) at time $T+\tau$, where $\tau$ denotes the interval of Time-to-Behavior (TTB). \emph{v}Beh-Pre enables the system to analyze the changes in driving scenes and make appropriate decisions. Because there are many safety-awareness factors in driving situations, we aim to accurately predict vehicle behavior while focusing on reasonable explanations for the model's prediction.

Current behavior prediction models for different road agents (e.g., pedestrians and vehicles) have entered the deep learning era \cite{Dang2017,Han2019,Hu2018}. However, the lack of interpretability still bothers this research field all the time. Especially for autonomous driving systems, every safe decision requires interpretability with much passion, as each accident caused by unreasonable and unjustified decisions has significant social influence \cite{autonomous-crash}. 

Decision Trees (DT) \cite{Quinlan1986,Murthy1994}  have the characteristic of hierarchically partitioning the sample space based on specific criteria and can provide users with hierarchical and interpretable decisions. However, the performance of decision trees is not satisfying compared with deep neural networks \cite{LeCun2015}. Therefore, some works have attempted to combine neural networks and decision trees to form the so-called ``Neural Decision Tree (NDT)"\cite{Li2022}, and apply NDT in some computer vision tasks \cite{Wan2020,Kim2022,Nauta2021}. Differently, we extend NDT by implanting the memory on the NDT with a behavior-type tree structure, which considers the \emph{episodic memory} constructed by historical vehicle locations and their scene interactions. It is called the Episodic Memory implanted Neural Decision Tree ({\tt\small \emph{e}Mem-NDT}). 
To fulfill a general tree structure, the construction process of {\tt\small \emph{e}Mem-NDT} is achieved by hierarchically clustering the text embedding of vehicle behavior descriptions. Different from the previous scene graph learning \cite{Lin2022,Suhail2021}, the behavior graph in this work can be simply constructed but owns a more understandable common sense of driving scenes. 
\begin{figure}[!t]
	\centering
	\includegraphics[width=\linewidth]{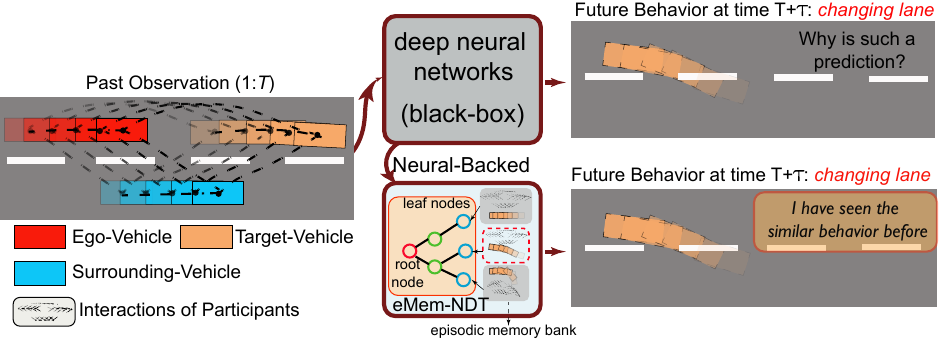}
	\caption{{\small{The \tt\small \emph{e}Mem-NDT} aims to retrospect the \emph{episodic memory} of certain kinds of vehicle behaviors and enhance the prediction trustworthiness.}}
	\label{fig1}	
	\vspace{-1.5em}
\end{figure}

Fig. \ref{fig1} illustrates the motivation of {\tt\small \emph{e}Mem-NDT} for the Beh-Pre. Specifically, in traditional deep learning-based approaches (such as Transformer \cite{Vaswani2017}, LSTM \cite{Hochreiter1997}, etc.), we can predict the future vehicle behavior label at time $T+\tau$ by encoding the historical locations and the interactions of vehicles. However, why is such a prediction? It cannot be explained. In this work, we leverage the retrospective memory function of humans \cite{Baddeley2013} to find the correlation between the prediction with the accumulated \emph{Episodic Memory Bank (EMB)} , containing the scene interactions and vehicle locations. Based on this, we implant this episodic memory into the NDT and make decisions by calculating the similarity between the current instance and the memory prototypes stored in the leaf node of {\tt\small \emph{e}Mem-NDT}.  The episodic memory bank in each leaf node stores the correlation between the behavior embedding of vehicles with their instances (consisting of raw vehicle locations and their interaction with road scenes) that can be obtained by baseline deep models. To reduce EMB redundancy, each memory prototype in EMB is filtered by ruling out over-similar instances. Then, by {\tt\small \emph{e}Mem-NDT}, we infer each instance in behavior prediction of vehicles by bottom-up \emph{Memory Prototype Matching (MPM)} (searching the appropriate leaf node and the links to the root node) and top-down \emph{Leaf Link Aggregation (LLA)} (obtaining the probability of future behaviors of vehicles for certain instances).  We believe that this inference process can find the appropriate memory prototype for each prediction. To summarize, the \textbf{contributions} are threefold.
\begin{itemize}
\item We construct an {\tt\small \emph{e}Mem-NDT}  to boost the interpretability of the vBeh-Pre, where every prediction can be given a reasonable interpretation by the memory behavior prototypes distilled from the training data. 
\item We implant the episodic memory in the leaf node of NDT to provide the introspective behavior prototypes of vehicles for each prediction, which also promotes the leaf node explainability. 
\item We infer the \emph{v}Beh-Pre by an efficient bottom-up link searching for each instance and obtain the behavior score at time $T+\tau$ by layer-wise link score aggregation. This formulation absorbs the vehicle behavior relations in certain driving scenes and improves the performance evaluated on the BLVD \cite{Xue2019} and LOKI \cite{Girase2021} datasets.

\end{itemize}

\section{Related Work}

\subsection{Vehicle Behavior Prediction}
Vehicle behavior prediction in this field focuses on Lane Changing (LC), Merging (M), Turning Left/Right (TL/TR), and Lane Keeping (LK) behaviors. 
Over decades, the deep learning-based formulations become the primary prototypes \cite{Dang2017,Tang2018,Tang2019,Izquierdo2019,Li2021,Mahajan2020,Han2019,Scheel2018,Hu2018,Li2023,Chen2022}. 

From the dynamic nature of vehicle behaviors, sequential networks, such as Long-Short-Term Memory (LSTM), are popular for modeling temporal correlations. For example, Dang \emph{et al.} \cite{Dang2017} employ the LSTM network to estimate the Time-To-Lane-Change (TTLC) by incorporating driver status, vehicle information (e.g., velocity), and environment clues (e.g., road lanes). Scheel \emph{et al.} \cite{Scheel2018} introduce the Bidirectional LSTM (BiLSTM) to make a cross-check for temporal correlation modeling the relation between vehicle trajectory points and LC intention. Compared with LSTM, BiLSTM obtains significant performance improvement. Mahajan \emph{et al.} \cite{Mahajan2020} propose a density clustering-based method to automatically identify LC and LK behavior patterns from unlabeled data, and further determined by the LSTM model. 

Besides the temporal information, the interaction among vehicles is also important for a collision-free behavior prediction. Recent works \cite{Li2023,Chen2022} model the social interaction of vehicles in driving scenes. They both use the multi-head attention mechanism in Transformer \cite{Vaswani2017} to model the interaction of vehicles. 

Deep learning models have shown excellent performance on vehicle behavior prediction, while they cannot explain the model's decisions. Most of these models focus on LK behavior, and the fine-grained but important behavior types, such as \emph{overtaking}, \emph{turning}, etc., are ignored. We believe that the \textbf{interpretable} model with \textbf{fine-grained behavior prediction} forms the basis of safe driving systems \cite{Tatsuya2022}. We introduce the Neural Decision Trees (NDTs) \cite{Li2022} to adapt to the large-scale data input with clear explainability. 

\subsection{Neural Decision Trees (NDTs)}
From the perspective of the tree structure of NDTs, it can be mainly divided into data-driven \cite{Tanno2019,Zhao2001} and pre-defined structures \cite{Frosst2017,Wan2020,Sethi1997}. Data-driven NDTs can grow based on the patterns within the data, but they are prone to overfitting and complex. For example, Tanno \emph{et al.} \cite{Tanno2019} propose a split gain-based approach that can start with a single node tree and dynamically grow or prune the decision tree based on the complexity and size of the data and learn in edges, path functions and leaf nodes. However, some common sense in driving scenes (e.g., traffic rules), can be used initially to design the structure of NDTs. Therefore, pre-defined tree structures are simple and easy to interpret with the scene knowledge prior. Frosst and Hinton \cite{Frosst2017} distill a trained neural network to a soft decision tree with a pre-defined structure for classification tasks. Inspired by this, Neural-Backed Decision Tree (NBDT) \cite{Wan2020} replaces the last linear layer of a convolutional neural network with a differentiable decision tree and a tree supervision loss to achieve the highly interpretable classification.

The interpretability ability of NDT nodes is different and can be divided into \emph{path-explainable} \cite{Wan2020,Tanno2019} and \emph{single-node explainable} \cite{Kim2022,Nauta2021} NDTs. 

The path-explainable NDT, such as NBDT \cite{Wan2020}, groups several single-step decisions from the root node to one leaf node and generates a hierarchical and sequential decision, owning the \emph{root-to-leaf} path decision explainability. Path-explainable NDT is easy to train but each single-step decision in one path is not explainable.
Contrarily, single-node explainable NDTs enable the interpretability for each node by learning the pattern prototype of each node representation. For example, Neural Prototype Tree (ProtoTree) \cite{Nauta2021} combines prototype learning with decision trees, and each decision step of one node can find the prototype learned from the dataset. However, single-node explainable NDTs require a complex training process. 
In the case of safe driving, we need a model to be lightweight, interpretable, and easily trained. Based on the neuropsychology of human retrospective memory \cite{Baddeley2013}, our work incorporates scene episodic memory into the NDT ({\tt\small \emph{e}Mem-NDT}), where the scene memory is obtained by pre-training a deep learning model for the prototype alignment between the feature embedding and each raw instance (\emph{i.e.}, vehicle locations and their interaction to other agents). The {\tt\small \emph{e}Mem-NDT} absorb the advantages of single-node explainable and path-explainable NDTs for vehicle behavior prediction.

\section{Method}

\textbf{Problem Formulation:} According to the definition in our previous work \cite{Xue2019,Fang2022}, Vehicle Behavior Prediction (\emph{v}Beh-Pre) refers to the process of estimating the future behaviors of Target Vehicles (TV). Given the historical state observation sequence of the $i^{th}$ vehicle $\mathcal{V}_i$=\{\textbf{v}$_i^1$,...,\textbf{v}$_i^t$,...,\textbf{v}$_i^T$\} up to time $T$, \emph{v}Beh-Pre is commonly formulated as a classification problem of behavior labels at time $T+\tau$, where $\tau$ is the Time-to-Behavior (TTB), and the vehicle state \textbf{v}$_i^t$ is represented by 
$\textbf{v}_i^t= [u_i^t, c_i^t, x_i^t, y_i^t, z_i^t, d_i^t]^T$ consisting of the vehicle ID $u_i^t$, vehicle label $c_i^t$, the 3D center point $(x_i^t, y_i^t, z_i^t)$, and the orientation $d_i^t$. 

In addition, this work further considers the interaction between the Target Vehicle (TV) and surrounding agents, such as pedestrians, other vehicles, etc. To represent the interaction information at time $t$, we introduce a graph $G^t_i(V_t, E_t)$ to represent the relationships of different road agents, where $V_t$ denotes the vertex at time $t$ represented by the states of different agents and the edge $E_t$ correlates the agent states at time $t$. Notably, the representation of the vertex at time $t$ is the same as the vehicle state \textbf{v}$_i^t$ while with more agent class $c_i^t$, \emph{i.e.}, involving pedestrians, cyclists, etc. 

\textbf{Episodic Memory:} For the observation with $T$ frames, this work defines episodic memory as the historical vehicle state $V_i$ and its interaction representation $\mathcal{G}_i=$\{$G^1_i$,..., $G^t_i$,..., $G^T_i$\} in $T$ window. Notably, the dynamic nature of driving scenes involves sporadically appearing or vanishing objects. Therefore, the numbers of vertex and edges in $G^t_i$ are different at different times.

With the denotation, the \emph{v}Beh-Pre is determined by maximizing the probability $\textbf{y}_i$ of behavior labels given the $\mathcal{V}_i$ and its interaction representation $\mathcal{G}_i$:
\begin{equation}
\hat{\textbf{y}}_i^{T+\tau}=\arg\max_{\textbf{y}_i} p(\textbf{y}_i|\mathcal{V}_i, \mathcal{G}_i),
\end{equation}
where $\hat{\textbf{y}}_i^{T+\tau}$ is the predicted one-hot vector to represent the behavior label (1-certain vehicle behavior).

\begin{figure}[!t]
	\centering
	\includegraphics[width=\linewidth]{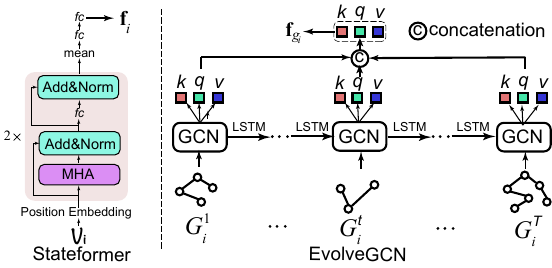}
	\caption{\small{The feature embedding for vehicle state and the dynamic interaction graphs by vehicle state transformer (Stateformer) \cite{Vaswani2017} and EvolveGCN, respectively.}}
	\vspace{-1em}
\end{figure}

\subsection{Base \emph{v}Beh-Pre Model}
For the problem in Eq. 1, there are a vast of sequential methods that can be utilized, such as various deep learning models, such as LSTM \cite{Hochreiter1997}, and Transformers \cite{Vaswani2017}. It is worth noting that, we encode the sequential vehicle states in a typical Transformer model \cite{Vaswani2017} with 2 layers of Multi-Head Attention (MHA), denoted as $\textbf{f}_i=\text{Trans}(\mathcal{V}_i)$. 

As for the interaction representation $\mathcal{G}_i$, because of the different number of agents at different times, this work introduces the EvolveGCN \cite{Pareja2020} to model the temporal correlation between the interaction graphs. EvolveGCN is constructed by Graph Convolutional Networks (GCN) \cite{Kipf2017} at different times and correlates the temporal relation by sequential networks, such as LSTM or GRU.  In our work, we further aggregate the embedding of GCN at different times with a self-attention model to model the temporal importance of interactions. Consequently, the feature embedding $\textbf{f}_{g^t_i}$ of the interaction graph representation at time $t$ is computed by involving the historical graph representations over $t$ times:
\begin{equation}
\textbf{f}_{g^t_i}=\text{EvolveGCN}(G^{1:t}_i),
\end{equation}
and
\begin{equation}\small
\textbf{f}_{g_i}=\text{ATT}(\text{concat}(\text{ATT}(\textbf{f}_{g^1_i}), ..., \text{ATT}(\textbf{f}_{g^t_i}),...,\text{ATT}(\textbf{f}_{g^T_i})))),
\end{equation}
where $\text{ATT}(.)$ is the self-attention operation with one head of query (q), key (k), and value (v).

The predicted label vector of vehicle behaviors  $\hat{\textbf{y}}_i^{T+\tau}$ at time $T+\tau$ of the $i^{th}$ vehicle is decoded by one fully-connected layer for the concatenation of $\textbf{f}_i$ and $\textbf{f}_{g_i}$:
\begin{equation}
\begin{aligned}
\textbf{g}_i=fc(\text{concat}(\textbf{f}_i,\textbf{f}_{g_i}),\theta), \\
\hat{\textbf{y}}_i^{T+\tau}=\delta(\textbf{g}_i),\verb'        '
\end{aligned}
\label{eq:4}
\end{equation}
where $fc$ is the fully-connected layer with the parameter $\theta$, and $\delta$ denotes the \emph{softmax} function.

It is clear that the base vBeh-Pre model has a black-box nature for interpreting the predictions. 

\subsection{Enhancing the Interpretability by {\tt\small \emph{e}Mem-NDT}}
In this work, we introduce the Neural Decision Trees (NDT) to interpret the prediction of vehicle behaviors by checking the decision process through hierarchical tree structures.  Inspired by the Neural-Backed DT (NBDT)\cite{Wan2020}, we take a backed grafting operation to replace the $softmax$ function $\delta(.)$ in Eq. \ref{eq:4} with a decision tree. Differently, we enforce the explainability of leaf nodes by implanting a kind of \textbf{Episodic Memory} that stores and retrospects the correlation between feature $\textbf{g}_i$ and the raw vehicle states and interaction graphs over $T$ frames.  
An element in eMem is denoted as $\{\textbf{g}_i:[\mathcal{V}_i,\mathcal{G}_i]\}$.

\begin{figure*}[!t]
	\centering
	\includegraphics[width=\linewidth]{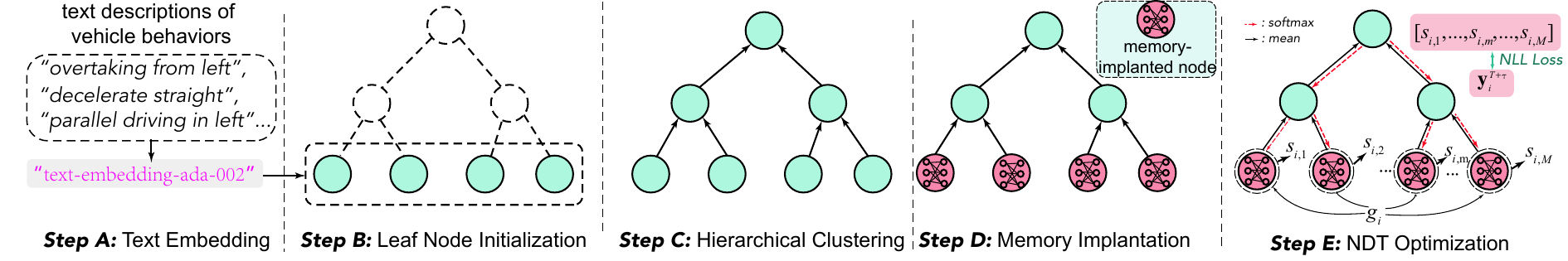}
	\caption{\small{The construction process of  {\tt\small \emph{e}Mem-NDT}, which contains five steps. Commonly, the number of leaf nodes is initialized by the vehicle behavior types. Hierarchical clustering groups the text embedding of leaf nodes and makes the tree grow. In particular, we implant the episodic memories of vehicles to the correlated leaf nodes, and the leaf nodes are further optimized by measuring the feature embedding of input instances (raw vehicle states and interactions) and stored episodic memories.}}
	\label{fig3}
\end{figure*}

\subsubsection{Construction of  {\tt\small \emph{e}Mem-NDT}}
Fig. \ref{fig3} illustrates the construction process of {\tt\small \emph{e}Mem-NDT}, which consists of five steps: \emph{behavior text embedding}, \emph{leaf node initialization},  \emph{hierarchical clustering}, \emph{memory implantation}, and \emph{NDT optimization}.

\textbf{Behavior Text Embedding for Leaf Node Initialization}: To boost the explainability, we encode the text descriptions of vehicle behaviors to initialize the leaf nodes of  {\tt\small \emph{e}Mem-NDT}, which leverage the behavior knowledge to fulfill an acceptable and reasonable tree structure. Specifically, we utilize the pre-trained Large-Language Model (LLM) {\tt\small text-embedding-ada-002} \cite{Neelakantan2022} to obtain the semantic vector of each behavior description. The number of leaf nodes of  {\tt\small \emph{e}Mem-NDT} equals the types of vehicle behaviors.

\textbf{Tree Growing by Hierarchical Clustering}: In this work, hierarchical clustering \cite{Jain1999} aims to aggregate the text embeddings of leaf nodes to high-level semantic concepts, such as \{``\emph{overtaking from left}", ``\emph{overtaking from right}"\} to ``\emph{overtaking}“, by computing the Euclidean distance of text embeddings. In the works, the growing of the tree is terminated until only one class (\emph{i.e.}, the ancestor node) is obtained in clustering.
It is worth noting that, besides the leaf nodes with clear semantic behavior descriptions, the inner nodes and the ancestor node are obtained by the feature clustering process. Actually, in the experiments, we take the GPT4 \cite{OpenAI2023} to label the semantic concepts for the inner and ancestor node, and the description presents a clear relation link between the certain instance and its inference path. 

\textbf{Memory Implantation}:  To facilitate the leaf node explainability, we implant the episodic memory in each leaf node. The episodic memory is constructed by introspecting all the vehicle states $\mathcal{V}_i$ ($i=1,..., N$) and the interaction graphs $\mathcal{G}_i$ over historical $T$ frames in the training data with $N$ instances. To make the memory representative, it is impossible to store all the instances in the memory bank. Inspired by the introspective memory \cite{Baddeley2013}, we design a Leaf Node Memory Filter (LNMF), which rules out the over-similar instance determined by a pre-defined threshold $\eta$ (evaluated in experiments) and maintains the representativeness of each memory prototype. The similarity of instances is computed by the \emph{cosine} function over the feature embeddings obtained by Eq. \ref{eq:4}. 
Algorithm \ref{alg1} illustrates the procedure of LNMF.

\begin{algorithm}[!t]\small
\textbf{Input:} \\
\hspace{2em} \( \mathcal{V}_i \): the vehicle state of \( i^{th} \) TV; \( \mathcal{G}_i \): interaction graphs \\
\hspace{2em} \( \eta \): Threshold for adding memories \\
\hspace{2em} \( v_{\text{max}} =-1 \): the similarity of instance features \\
\textbf{Forward:} \\
\hspace{2em} \(\triangleright\) Initialize memory bank \( \mathcal{M} = \{\textbf{E}_m = \emptyset, \textbf{R}_m = \emptyset\} \) \\
\hspace{2em} \( \textbf{E}_m \): Feature memory bank for the $m^{th}$ leaf node\\
\hspace{2em} \( \textbf{R}_m \): Interaction memory bank for the $m^{th}$ leaf node\\
\hspace{2em} \(\triangleright\) Extract instance feature  by Eq. \ref{eq:4} \\
\hspace{2em} \textbf{if} \( \textbf{E}_m \) is empty: \\
\hspace{3em} \( \textbf{g}_i \rightarrow \textbf{E}_m \), $[\mathcal{V}_i,\mathcal{G}_i] \rightarrow \textbf{R}_m$ \\
\hspace{2em} \textbf{else}: \\
\hspace{3em} \textbf{for} \( \textbf{e} \) in \( \textbf{E}_m \): \\
\hspace{4em}  \( v = \cos \langle \textbf{e}, \textbf{g}_i \rangle \) \\
\hspace{4em} \textbf{if} \( v \geq v_{\text{max}} \): \\
\hspace{5em} \( v_{\text{max}} = v \) \\
\hspace{3em} \textbf{if} \( v_{\text{max}} \leq \eta \): \\
\hspace{3em} \( \textbf{g}_i \rightarrow \textbf{E}_m \), \( (\mathcal{V}_i, \mathcal{G}_i) \rightarrow \textbf{R}_m \) \\
\textbf{Output:} $\textbf{E}_m$, $\textbf{R}_m$ \\
\caption{\small{Leaf Node Memory Filtering (LNMF)}}
\label{alg1}
\end{algorithm}

Assume we obtain $K$ memory bases in the $m^{th}$ leaf node after memory filtering for all the training instances, which means that for $m^{th}$ type of vehicle behavior, we have $K$ episodic memory bases. Because of the sample imbalance of different vehicle behaviors, we can obtain different numbers of memory bases for each leaf node, \emph{i.e.}, $\mathcal{K}=[K_1,..., K_m,..., K_M]$ for $M$ leaf nodes.

\textbf{NDT Optimization}: Different from traditional decision trees, neural decision trees implant a neural network into some tree nodes. To form a trainable neural decision tree, we further implant two fully connected layers ($fc(.,.)$) in each leaf node, which can be treated as a transforming network to project the feature vector $\textbf{g}_i$ obtained by the Eq. \ref{eq:4} and the feature embedding in the memory bank $\textbf{E}_m$ of $m^{th}$ leaf node to a shared feature space. With that, the correlation between the output of the base vBeh-Pre model and the episodic memory can be computed. Therefore, the correlation for the $i^{th}$ instance input is computed by:
\begin{equation}
\gamma_{i,m}=\rho \cdot cos(\mathcal{H}(\textbf{g}_i),\mathcal{H}(\textbf{e}_k)), \textbf{e}_k\in \textbf{E}_m, k=1,...,K_m,
\label{eq:5}
\end{equation}
where $\mathcal{H}(\textbf{a})=fc(fc(\textbf{a},\theta_1),\theta_2)$, $\theta_1$ and $\theta_2$ are the parameters of the fully connected layers, and $\rho>1$ is a hyperparameter to amplify the cosine value for avoiding the gradient vanishing issue owning to that the absolute value of cosine value is less than 1. 

In order to optimize the $\mathcal{H}(.)$, we model a \emph{bottom-up Memory Prototype Matching (MPM)} and a \emph{top-down Leaf Link Aggregation (LLA)} by inputting the feature embedding of all instances to traverse the NDT structure. In the traversing process, when arriving at the leaf nodes, we use Eq. \ref{eq:5} to compute the similarity score $\gamma_{i,m}$ between the input feature vector and the filtered instance feature vectors. Then, the bottom-up MPM from the leaf nodes to the up layer  (visualized as the black solid lines in \textbf{Step E} in Fig. \ref{fig3}) is fulfilled by:
\begin{equation}
\gamma_{i,M+z}=\text{mean}(\gamma_{i,[m\rightarrow (M+z)]}), 
\end{equation}
where $\gamma_{i, M+z}$ is the similarity score of the parent nodes with the links to its child leaf nodes $[m\rightarrow (M+z)]$. Like this, bottom-up MPM traverses all the layers until the ancestor node is reached. Then, the top-down LLA begins by a layer-wise classifier of \emph{softmax} in NDT (denoted as the red dashed line in \textbf{Step E} in Fig. \ref{fig3}), and the classifier from the penultimate layer to the leaf nodes is defined as:
\begin{equation}
s_{i,m}=\text{softmax}(\gamma_{i,[{(M+z)\rightarrow m}]})
\end{equation} 
where $s_{i,m}$ is the behavior label probability in the $m^{th}$ leaf node of {\tt\small \emph{e}Mem-NDT}. Consequently, we obtain a probability vector $S_i=[s_{i,1},..., s_{i,m},...,s_{i,M}]$ at the leaf node layer. To optimize $\mathcal{H}()$, we introduce the Negative Log-Likelihood (NLL) loss to match the probability vector with the ground-truth behavior label $y$, denoted as:
\begin{equation}\small
 \mathcal{L}(S, \textbf{Y}^{T+\tau}) = \frac{1}{N^b} \sum_{i=1}^{N^b} L_i(S_i, \textbf{y}_i^{T+\tau}) = \frac{1}{N^b} \sum_{i=1}^{N^b} (-\log(s_{i,id})) 
\end{equation}
where $\textbf{Y}^{T+\tau}$ is the label set in one batch with $N^b$ samples, $\textbf{y}_i^{T+\tau}$ is a one-hot vector to represent the ground-truth behavior label ($id$==1 means the active vehicle behavior).
 
\subsubsection{Testing Inference}

Here we use soft decision as our decision-making mechanism. Assume we have $J$ testing instances that contain the vehicle state $\mathcal{V}_j$ and the interaction graphs $\mathcal{G}_j$ with the same $T$ time step observation. We first use Eq. \ref{eq:4} to obtain the feature embedding $\textbf{g}_j$. Then, $\textbf{g}_j$ is fed into the trained {\tt\small \emph{e}Mem-NDT} through a soft decision-making process, \emph{i.e.}, the \textbf{bottom-up MPM} by Eq. 6 and \textbf{top-down LLA}  Eq. 7. Then the behavior label probability vector $\textbf{s}_j\in \mathbb{R}^{1\times M}$ on all the leaf nodes is calculated by the accumulating product of probabilities of root-to-leaf paths, and the final behavior label probability of $j^{th}$ instance is:
\begin{equation}
y_j^{T+\tau}=\arg\mathrm{max}_{m}\textbf{s}_j(m).
\end{equation}

\section{ Experiments}
\subsection{Datasets and Metrics}
Based on the investigation \cite{Fang2022}, BLVD \cite{Xue2019} and LOKI \cite{Girase2021} datasets have the most fine-grained vehicle behaviors with 3D LiDAR data. They are described as follows. 

\textbf{BLVD}: The BLVD \cite{Xue2019} dataset is collected in Changshu City, Jiangsu Province, China. It contains different driving scenarios with various light conditions and 13 different vehicle behavior labels under the Bird's Eye View (BEV) observation. BLVD consists of 77,434 vehicle instances (with a training and testing ratio of 8:2) and a frame rate of 10Hz, where each instance owns 0.5s observation and 1s prediction.

 \textbf{LOKI:} The LOKI \cite{Girase2021} dataset contains a variety of complex traffic situations and pedestrian behaviors and is primarily used for trajectory prediction. LOKI was constructed by Honda American Institute, Honda R\&D Center, and the University of California, Berkeley, USA. Compared with BLVD, LOKI has 9 types of vehicle behaviors. The frame rate of LOKI is 5Hz and contains 7,491 vehicle instances (with the same training and testing ratio of 8:2) with 0.6s observation and 1s prediction. Fig. 4 presents the instance distribution of the vehicle behavior categories. The LOKI has a rather smaller sample scale and a more severe sample imbalance issue than BLVD.
 
\textbf{Metrics:}
For the behavior prediction, similar to other works \cite{Zhang2020}, we adopt the Precision, Recall, and F1-score ($\beta=1$) \cite{Xue2019} as the evaluation metrics.
 \begin{figure}[!t]
	\centering
	\includegraphics[width=\linewidth]{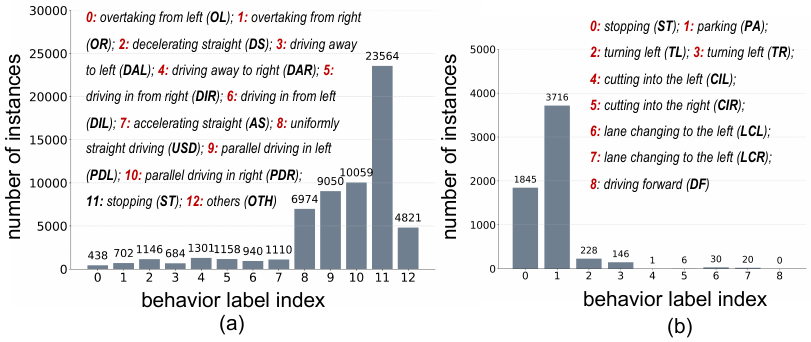}
	\caption{\small{The instance distribution of (a) BLVD dataset and (b) LOKI dataset. LOKI has a more severe imbalance issue than BLVD.}}
	\label{fig4}	
\end{figure}

\subsection{Implementation Details}
First, we train the ``base vBeh-Pre" model and then use the trained ``base vBeh-Pre" as the feature extractor in the subsequent process. Then, the NDT optimization is trained after fixing the ``base vBeh-Pre" model. \ding{182} For the ``base vBeh-Pre" model, we use the Cross-Entropy loss for training with 80 epochs. The weight decay of the Adam optimizer is $1 \times 10^{-5}$ and the learning rates are $0.0005$ and $0.005$ respectively for the Stateformer and EvolveGCN. The batch size $N^b$ in Eq. 8 is set as 64.  \ding{183} After training the ``base vBeh-Pre" model, we use the Negative Log-Likelihood (NLL) loss to supervise the neural tree learning with 5 epochs (converged). Here, we set the learning rate to $5 \times 10^{-5}$ and the weight decay to $1\times 10^{-6}$. We set $\rho$ in Eq. 5 as $30$ in this work. 
\begin{table}[!t]\small
    \centering
    \caption{\small{Performance comparison on the \textbf{BLVD/LOKI} dataset with $\tau=1$. The best values are marked by \textbf{bold} font. }}
        \renewcommand{\arraystretch}{1.2}
    \begin{tabular}{c|ccc}
\toprule[0.8pt]
    Model & Precision & Recall & F1-score \\
    \hline
LSTM \cite{Hochreiter1997} &0.75/\textbf{0.34}&0.88/0.26&0.82/0.30\\
    Transformer \cite{Vaswani2017} & 0.68/0.26 & 0.89/0.23 & 0.79/0.24 \\
    CNN\_BiLSTM \cite{Zhang2020}  & 0.81/0.33 & 0.89/0.26 & 0.85/0.29 \\
    \hline
    NBDT \cite{Wan2020}  & 0.80/0.30 & \textbf{0.94}/0.32 & 0.86/0.29 \\
    \hline
    vBeh-Pre &0.92/0.30&0.91/0.34&0.91/0.32\\ 
eMem-NDT & \textbf{0.94}/0.33 &0.92/\textbf{0.37} & \textbf{0.93}/\textbf{0.33} \\
\toprule[0.8pt]
    \end{tabular}
    \label{tab1}
\end{table}

\subsection{Overall Evaluations}
To the best of our knowledge, there are few works \cite{Dang2017,Tang2018,Tang2019,Izquierdo2019,Li2021,Mahajan2020,Han2019,Scheel2018,Hu2018} on predicting vehicle behaviors, but most of them focus on predicting vehicle lane changing behavior, rarely involve the fine-grained vehicle behavior prediction. Here, we choose CNN\_BiLSTM \cite{Zhang2020}, LSTM \cite{Hochreiter1997}, and the typical Transformer \cite{Vaswani2017} in the comparison. Besides, we also take the NBDT \cite{Wan2020} with the official parameter setting in the evaluation comparison by a neural-backed decision tree after replacing the fully connected layer in Eq. 4. Here we set $\tau=1s$ on the BLVD dataset and LOKI datasets. Table. \ref{tab1} shows the Precision, Recall, and F1-score in the comparison. From the results, compared with traditional deep learning methods, we can see that our methods, including the vBeh-Pre and the eMem-NDT, generate better Precision and F1 values on the BLVD and LOKI datasets. However, because of the severe sample imbalance issue, the metric values of the LOKI dataset have a large scale to be improved. Some vehicle behaviors in LOKI only have less than 30 samples. Contrarily, LSTM performs a competitive performance (with the F1 value of 0.30) with the simplest model architecture.

\subsection{Generalization for Few-Shot Behaviors}
As shown in Fig. 4, the first eight types of vehicle behaviors in the BLVD dataset have rather fewer samples than other ones. Therefore, we check the method performance on these few-shot samples. Table. \ref{tab2} presents the F1 values on different vehicle behaviors. From these results, we can see that the base vBeh-Pre model provides a promising baseline. Almost all the behaviors obtain F1 values greater than 0.82, and our eMem-NDT obtains better performance than NBDT and the base vBeh-Pre model. Among them, the ``\emph{accelerating straight} (AS)" behavior obtains worse results than other behaviors. With episodic memory, our eMem-NDT is significantly better than NBDT \cite{Wan2020}, which verifies the promotion role of the memory mechanism in NDT.

\begin{table}[t]\small
    \centering
    \caption{\small{F1 value of base vBeh-Pre, NDBT, and our eMem-NDT on the few-shot behaviors in the BLVD dataset.}}
        \renewcommand{\arraystretch}{1.2}
            \setlength{\tabcolsep}{0.8mm}{
    \begin{tabular}{c|cccccccc}
\toprule[0.8pt]
     Model&OL&OR  &DS  &DAL &DAR&DIR&DIL&AS \\
    \hline
      NBDT  \cite{Wan2020} &0.82&0.81&0.82&0.93&0.88&0.86&0.82&0.66\\
     vBeh-Pre&0.89&0.87&0.91&0.95&\textbf{0.94}&0.90&0.89&0.82\\
eMem-NDT&\textbf{0.94}&\textbf{0.93}&\textbf{0.93}&\textbf{0.96}&\textbf{0.94}&\textbf{0.92}&\textbf{0.92}& \textbf{0.87}\\
\toprule[0.8pt]
    \end{tabular}}
    \vspace{-1em}
    \label{tab2}
\end{table}
\begin{figure}[!t]
\begin{minipage}[b]{0.55\linewidth}
\centering
\includegraphics[width=\linewidth]{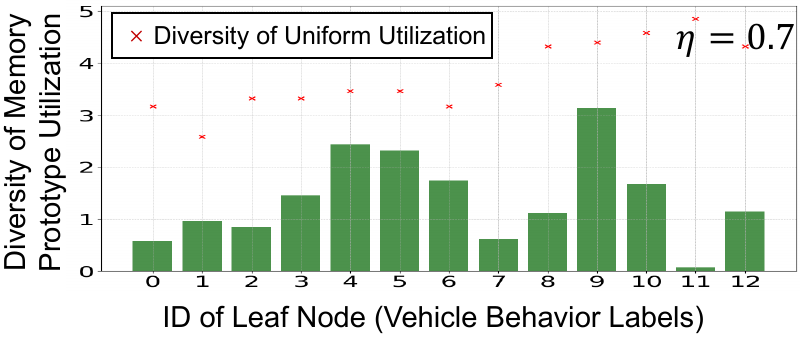}
\caption{\small{The diversity of memory prototype utilization ($\eta=0.7$).}}
\label{fig5}
\vspace{-3em}
\end{minipage}
\begin{minipage}[b]{0.38\linewidth}
\centering
\footnotesize
 \renewcommand{\arraystretch}{1.2}
 \setlength{\tabcolsep}{0.3mm}{
 \centering
   \captionof{table}{\footnotesize{Performance, w.r.t., $\eta$.}}}
  \label{tab3}
\begin{tabular}{l|cccc}
\toprule[0.8pt]
 $\eta$& $\text{Pre}\uparrow$ & $\text{Rec}\uparrow$& $\text{F1}\uparrow$&\text{EMB} \\
 \hline
0.3& 0.90&0.89&0.89&26\\
0.7& 0.92 & 0.92 & 0.92 &   95 \\
0.8& 0.93 & 0.91 & 0.92 &    192    \\
0.9& \textbf{0.94} & 0.92 & 0.93&509\\   
\toprule[0.8pt]
\end{tabular}
\end{minipage}
\vspace{-0.8em}
\end{figure}

\begin{figure}[!t]
	\centering
	\includegraphics[width=\linewidth]{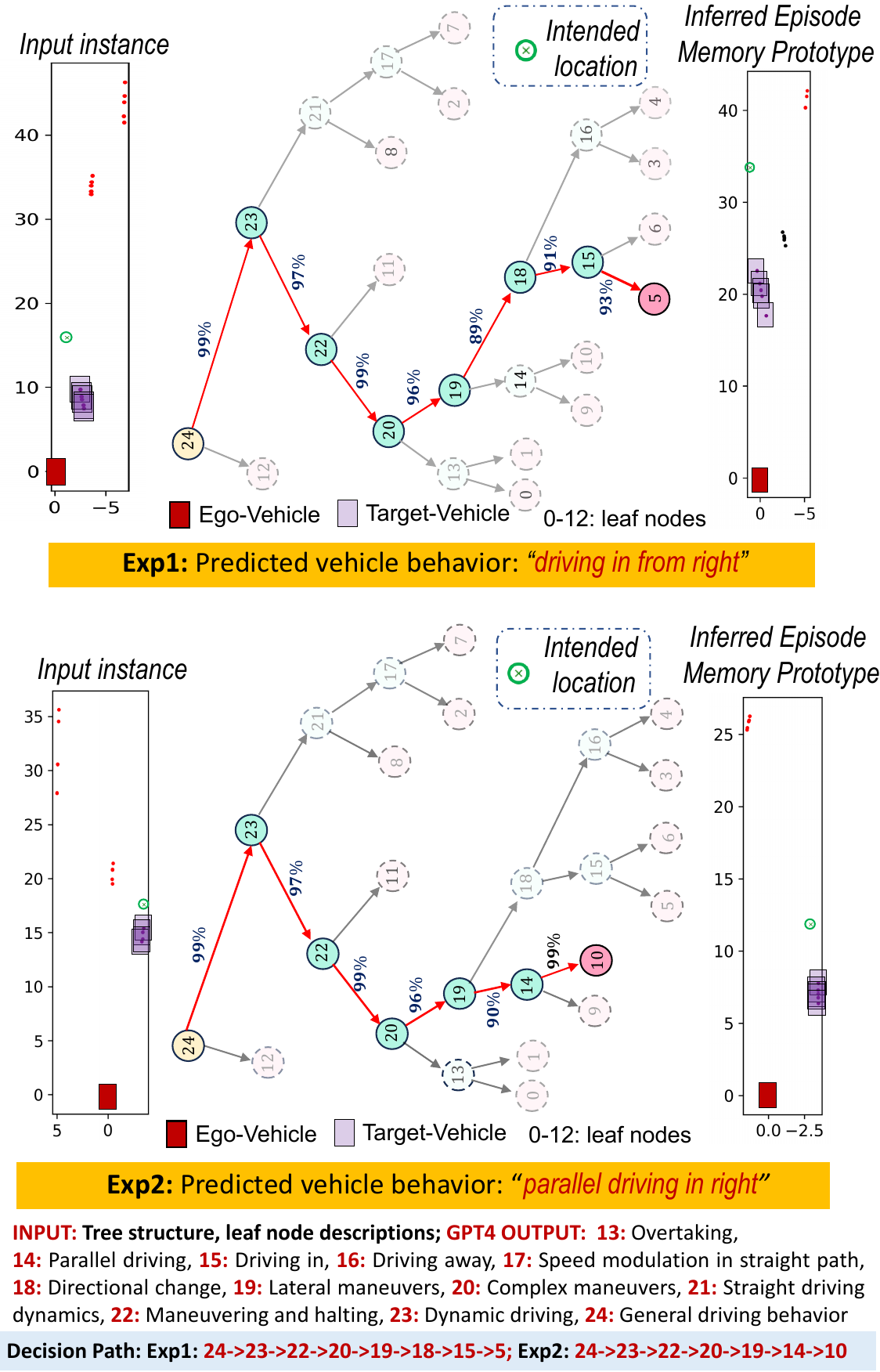}
	\caption{\small{The inference process for one input instance on the testing set of BLVD, where the episodic memory prototype of the behavior of the target vehicle is retrospected and matched. }}
	\label{fig6}	
	\vspace{-1em}
\end{figure}

\subsection{The Utilization Sparsity in Episodic Memory Bank (EMB)}
Actually, for one vehicle behavior instance, the utilization of memory prototypes of NDT leaf nodes has the sparsity, \emph{i.e.}, only a small proportion of memory prototypes stored in one leaf node of eMem-NDT play the role in the prediction. Fig. \ref{fig5} demonstrates the diversity of memory prototype utilization computed by the entropy of the utilization frequencies of all prototypes of one leaf node. In other words, if we set a smaller threshold $\eta$ to filter the Episodic Memory Bank (EMB), \emph{i.e.}, to generate fewer memory prototypes, the prototype utilization may remain stable, and the performance may be not vulnerable to the $\eta$, verified by Table. III.
However, with episodic memory implantation, the prediction of vehicle prediction is more explainable. Fig. \ref{fig6} illustrates the inference process of two instances, where the structure of NDT is obtained by hierarchically clustering the text descriptions of vehicle behaviors. Given the input instance, we can obtain a clear \emph{root-to-leaf} path with the correlated memory prototype. From Fig. \ref{fig6}, the inferred episodic memory prototype of the vehicle behavior is very similar to the input instance.

 We also label the semantic concept of other nodes in {\tt\small \emph{e}Mem-NDT} by GPT4 API \cite{OpenAI2023} given the text descriptions of leaf nodes and the NDT structure with a well-designed prompt\footnote{\scriptsize{``You are an expert in driving behavior classification. Goal: Assist in the task of driving behavior prediction. Task: 1. Based on the clustering structure, assign abstract semantics and explanations to intermediate nodes giving the semantic descriptions of the leaf nodes. Note - Decision-making must be rigorous and accurate."}}. From the inference path, the semantic labels gradually become fine-grained and reasonable for the prediction.

\section{Conclusions}
This work presents a new way to interpret the fine-grained prediction of vehicle behaviors by implanting a kind of episodic memory (consisting of vehicle locations and their scene interactions) into a Neural Decision Tree ({\tt\small \emph{e}Mem-NDT}), which is trainable and fulfilled by a neural-backed operation to replace the \emph{softmax} layer in a base model. The structure of {\tt\small \emph{e}Mem-NDT} is constructed by text embedding of vehicle behavior description, which is simple and has a clear common sense of driving scenes. From the results, {\tt\small \emph{e}Mem-NDT} can provide a clear inference process for each prediction, and improve the performance on the fine-grained vehicle behavior prediction. In the future, we will explore the episodic memory implanted in NDT to interpret the joint prediction of vehicle trajectories and behaviors.

\bibliographystyle{IEEEtran}
\bibliography{root1}

\begin{thebibliography}{10}
\providecommand{\url}[1]{#1}
\csname url@rmstyle\endcsname
\providecommand{\newblock}{\relax}
\providecommand{\bibinfo}[2]{#2}
\providecommand\BIBentrySTDinterwordspacing{\spaceskip=0pt\relax}
\providecommand\BIBentryALTinterwordstretchfactor{4}
\providecommand\BIBentryALTinterwordspacing{\spaceskip=\fontdimen2\font plus
\BIBentryALTinterwordstretchfactor\fontdimen3\font minus
  \fontdimen4\font\relax}
\providecommand\BIBforeignlanguage[2]{{%
\expandafter\ifx\csname l@#1\endcsname\relax
\typeout{** WARNING: IEEEtran.bst: No hyphenation pattern has been}%
\typeout{** loaded for the language `#1'. Using the pattern for}%
\typeout{** the default language instead.}%
\else
\language=\csname l@#1\endcsname
\fi
#2}}

\bibitem{Dang2017}
H.~Q. Dang, J.~F{\"{u}}rnkranz, A.~Biedermann, and M.~H{\"{o}}pfl,
  ``Time-to-lane-change prediction with deep learning,'' in \emph{ITSC}, 2017,
  pp. 1--7.

\bibitem{Han2019}
T.~Han, J.~Jing, and {\"{U}}.~{\"{O}}zg{\"{u}}ner, ``Driving intention
  recognition and lane change prediction on the highway,'' in \emph{IV}, 2019,
  pp. 957--962.

\bibitem{Hu2018}
Y.~Hu, W.~Zhan, and M.~Tomizuka, ``Probabilistic prediction of vehicle semantic
  intention and motion,'' in \emph{IV}, 2018, pp. 307--313.

\bibitem{autonomous-crash}
T.~Ampe, ``Autonomous accidents: The ethics of self-driving car crashes,''
  \url{https://vce.usc.edu/volume-4-issue-2/autonomous-accidents-the-ethics-of-self-driving-car-crashes/},
  2020.

\bibitem{Quinlan1986}
J.~R. Quinlan, ``Induction of decision trees,'' \emph{Mach. Learn.}, vol.~1,
  no.~1, pp. 81--106, 1986.

\bibitem{Murthy1994}
S.~K. Murthy, S.~Kasif, and S.~Salzberg, ``A system for induction of oblique
  decision trees,'' \emph{J. Artif. Intell. Res.}, vol.~2, pp. 1--32, 1994.

\bibitem{LeCun2015}
Y.~LeCun, Y.~Bengio, and G.~Hinton, ``Deep learning,'' \emph{Nature}, vol. 521,
  no. 7553, pp. 436--444, 2015.

\bibitem{Li2022}
H.~Li, J.~Song, M.~Xue, H.~Zhang, J.~Ye, L.~Cheng, and M.~Song, ``A survey of
  neural trees,'' \emph{CoRR}, vol. abs/2209.03415, 2022.

\bibitem{Wan2020}
A.~Wan, L.~Dunlap, D.~Ho, J.~Yin, S.~Lee, H.~Jin, S.~Petryk, S.~A. Bargal, and
  J.~E. Gonzalez, ``{NBDT:} neural-backed decision trees,'' \emph{CoRR}, vol.
  abs/2004.00221, 2020.

\bibitem{Kim2022}
S.~Kim, J.~Nam, and B.~Ko, ``Vit-net: Interpretable vision transformers with
  neural tree decoder,'' in \emph{ICML}, 2022, pp. 11\,162--11\,172.

\bibitem{Nauta2021}
M.~Nauta, R.~van Bree, and C.~Seifert, ``Neural prototype trees for
  interpretable fine-grained image recognition,'' in \emph{CVPR}, 2021, pp.
  14\,933--14\,943.

\bibitem{Lin2022}
X.~Lin, C.~Ding, Y.~Zhan, Z.~Li, and D.~Tao, ``Hl-net: Heterophily learning
  network for scene graph generation,'' in \emph{CVPR}, 2022, pp.
  19\,454--19\,463.

\bibitem{Suhail2021}
M.~Suhail, A.~Mittal, B.~Siddiquie, C.~Broaddus, J.~Eledath, G.~G. Medioni, and
  L.~Sigal, ``Energy-based learning for scene graph generation,'' in
  \emph{CVPR}, 2021, pp. 13\,936--13\,945.

\bibitem{Vaswani2017}
A.~Vaswani, N.~Shazeer, N.~Parmar, J.~Uszkoreit, L.~Jones, A.~N. Gomez,
  L.~Kaiser, and I.~Polosukhin, ``Attention is all you need,'' in \emph{NIPS},
  2017, pp. 5998--6008.

\bibitem{Hochreiter1997}
S.~Hochreiter and J.~Schmidhuber, ``Long short-term memory,'' \emph{Neural
  Comput.}, vol.~9, no.~8, pp. 1735--1780, 1997.

\bibitem{Baddeley2013}
A.~Baddeley, ``Essentials of human memory (classic ed.),'' 2013.

\bibitem{Xue2019}
J.~Xue, J.~Fang, T.~Li, B.~Zhang, P.~Zhang, Z.~Ye, and J.~Dou, ``{BLVD:}
  building {A} large-scale 5d semantics benchmark for autonomous driving,'' in
  \emph{ICRA}, 2019, pp. 6685--6691.

\bibitem{Girase2021}
H.~Girase, H.~Gang, S.~Malla, J.~Li, A.~Kanehara, K.~Mangalam, and C.~Choi,
  ``{LOKI:} long term and key intentions for trajectory prediction,'' in
  \emph{ICCV}, 2021, pp. 9783--9792.

\bibitem{Tang2018}
J.~Tang, F.~Liu, W.~Zhang, R.~Ke, and Y.~Zou, ``Lane-changes prediction based
  on adaptive fuzzy neural network,'' \emph{Expert Systems with Applications},
  vol.~91, pp. 452--463, 2018.

\bibitem{Tang2019}
J.~Tang, S.~Yu, F.~Liu, X.~Chen, and H.~Huang, ``A hierarchical prediction
  model for lane-changes based on combination of fuzzy c-means and adaptive
  neural network,'' \emph{Expert Systems with Applications}, vol. 130, pp.
  265--275, 2019.

\bibitem{Izquierdo2019}
R.~Izquierdo, {\'{A}}.~Quintanar, I.~Parra, D.~F. Llorca, and M.~{\'{A}}.
  Sotelo, ``Experimental validation of lane-change intention prediction
  methodologies based on {CNN} and {LSTM},'' in \emph{ITSC}, 2019, pp.
  3657--3662.

\bibitem{Li2021}
L.~Li, W.~Zhao, C.~Xu, C.~Wang, Q.~Chen, and S.~Dai, ``Lane-change intention
  inference based on {RNN} for autonomous driving on highways,'' \emph{{IEEE}
  Trans. Veh. Technol.}, vol.~70, no.~6, pp. 5499--5510, 2021.

\bibitem{Mahajan2020}
V.~Mahajan, C.~Katrakazas, and C.~Antoniou, ``Prediction of lane-changing
  maneuvers with automatic labeling and deep learning,'' \emph{Transportation
  Research Record}, vol. 2674, pp. 336 -- 347, 2020.

\bibitem{Scheel2018}
O.~Scheel, L.~A. Schwarz, N.~Navab, and F.~Tombari, ``Situation assessment for
  planning lane changes: Combining recurrent models and prediction,'' in
  \emph{ICRA}, 2018, pp. 2082--2088.

\bibitem{Li2023}
Z.~Li, Y.~Wang, and Z.~Zuo, ``Interaction-aware prediction for cut-in
  trajectories with limited observable neighboring vehicles,'' \emph{{IEEE}
  Trans. Intell. Veh.}, vol.~8, no.~3, pp. 2148--2161, 2023.

\bibitem{Chen2022}
X.~Chen, H.~Zhang, F.~Zhao, Y.~Hu, C.~Tan, and J.~Yang, ``Intention-aware
  vehicle trajectory prediction based on spatial-temporal dynamic attention
  network for internet of vehicles,'' \emph{{IEEE} Trans. Intell. Transp.
  Syst.}, vol.~23, no.~10, pp. 19\,471--19\,483, 2022.

\bibitem{Tatsuya2022}
T.~Sakai and T.~Nagai, ``Explainable autonomous robots: a survey and
  perspective,'' \emph{Advanced Robotics}, vol.~36, no. 5-5, pp. 219--238,
  2022.

\bibitem{Tanno2019}
R.~Tanno, K.~Arulkumaran, D.~C. Alexander, A.~Criminisi, and A.~V. Nori,
  ``Adaptive neural trees,'' in \emph{ICML}, 2019, pp. 6166--6175.

\bibitem{Zhao2001}
Q.~Zhao, ``Evolutionary design of neural network tree-integration of decision
  tree, neural network and {GA},'' in \emph{CEC}, 2001, pp. 240--244.

\bibitem{Frosst2017}
N.~Frosst and G.~E. Hinton, ``Distilling a neural network into a soft decision
  tree,'' in \emph{AI*IA}, 2017.

\bibitem{Sethi1997}
I.~K. Sethi and J.~H. Yoo, ``Structure-driven induction of decision tree
  classifiers through neural learning,'' \emph{Pattern Recognit.}, vol.~30,
  no.~11, pp. 1893--1904, 1997.

\bibitem{Fang2022}
J.~Fang, F.~Wang, P.~Shen, Z.~Zheng, J.~Xue, and T.~Chua, ``Behavioral
  intention prediction in driving scenes: {A} survey,'' \emph{CoRR}, vol.
  abs/2211.00385, 2022.

\bibitem{Pareja2020}
A.~Pareja, G.~Domeniconi, J.~Chen, T.~Ma, T.~Suzumura, H.~Kanezashi, T.~Kaler,
  T.~B. Schardl, and C.~E. Leiserson, ``Evolvegcn: Evolving graph convolutional
  networks for dynamic graphs,'' in \emph{AAAI}, 2020, pp. 5363--5370.

\bibitem{Kipf2017}
T.~N. Kipf and M.~Welling, ``Semi-supervised classification with graph
  convolutional networks,'' in \emph{ICLR}, 2017.

\bibitem{Neelakantan2022}
A.~Neelakantan, T.~Xu, R.~Puri, A.~Radford, J.~M. Han, J.~Tworek, Q.~Yuan,
  N.~Tezak, J.~W. Kim, C.~Hallacy, J.~Heidecke, P.~Shyam, B.~Power, T.~E.
  Nekoul, G.~Sastry, G.~Krueger, D.~Schnurr, F.~P. Such, K.~Hsu, M.~Thompson,
  T.~Khan, T.~Sherbakov, J.~Jang, P.~Welinder, and L.~Weng, ``Text and code
  embeddings by contrastive pre-training,'' \emph{CoRR}, vol. abs/2201.10005,
  2022.

\bibitem{Jain1999}
A.~K. Jain, M.~N. Murty, and P.~J. Flynn, ``Data clustering: {A} review,''
  \emph{{ACM} Comput. Surv.}, vol.~31, no.~3, pp. 264--323, 1999.

\bibitem{OpenAI2023}
OpenAI, ``{GPT-4} technical report,'' \emph{CoRR}, vol. abs/2303.08774, 2023.

\bibitem{Zhang2020}
H.~Zhang, Z.~Nan, T.~Yang, Y.~Liu, and N.~Zheng, ``A driving behavior
  recognition model with bi-lstm and multi-scale {CNN},'' in \emph{IV}, 2020,
  pp. 284--289.

\end{thebibliography}

\end{document}